\documentclass[sigconf]{acmart}
\usepackage{graphicx} 
\usepackage{subfigure}
\usepackage{balance}
\usepackage{multirow}
\usepackage{bbm}
\usepackage{color}
\usepackage{epstopdf}
\usepackage{hyperref}

\pdfoutput=1
\AtBeginDocument{%
  }

\copyrightyear{2023} 
\acmYear{2023} 
\setcopyright{acmlicensed}
\acmConference[MM '23]{Proceedings of the 31st ACM International Conference on Multimedia}{October 29-November 3, 2023}{Ottawa, ON, Canada}
\acmBooktitle{Proceedings of the 31st ACM International Conference on Multimedia (MM '23), October 29-November 3, 2023, Ottawa, ON, Canada}
\acmPrice{15.00}
\acmDOI{10.1145/3581783.3612317}
\acmISBN{979-8-4007-0108-5/23/10}

\begin{document}
% \fancyhead{}
\title{Benign Shortcut for Debiasing: Fair Visual Recognition via Intervention with Shortcut Features}


\author{Yi Zhang}
\authornote{This work was done when the author interned at Peng Cheng Lab.}
\affiliation{%
  \institution{School of Computer and Information Technology \& Beijing Key Lab of Traffic Data Analysis and Mining, Beijing Jiaotong University, China}
%   \institution{$^{2}$Peng Cheng Laboratory, China}
  \state{}
  \country{}}
\email{yi.zhang@bjtu.edu.cn}

\author{Jitao Sang}
\authornote{Corresponding authors}
\affiliation{%
  \institution{$^{1}$School of Computer and Information Technology \& Beijing Key Lab of Traffic Data Analysis and Mining, Beijing Jiaotong University, China}
  \institution{$^{2}$Peng Cheng Lab, Shenzhen, China}
  \state{}
  \country{}}
\email{jtsang@bjtu.edu.cn}

\author{Junyang Wang}
\affiliation{%
  \institution{School of Computer and Information Technology \& Beijing Key Lab of Traffic Data Analysis and Mining, Beijing Jiaotong University, China}
%   \institution{$^{2}$Peng Cheng Laboratory, China}
  \state{}
  \country{}}
\email{21120406@bjtu.edu.cn}

\author{Dongmei Jiang}
\affiliation{%
  \institution{Peng Cheng Lab, Shenzhen, China}
  \state{}
  \country{}}
\email{jiangdm@pcl.ac.cn }

\author{Yaowei Wang}
\affiliation{%
  \institution{Peng Cheng Lab, Shenzhen, China}
  \state{}
  \country{}}
\email{wangyw@pcl.ac.cn}

% \author{Yi Zhang}
% \affiliation{%
%   \institution{Beijing Jiaotong University}
%   \state{Beijing}
% \email{yi.zhang@bjtu.edu.cn}

% \author{Junyang Wang}
% \affiliation{%
%   \institution{Beijing Jiaotong University}
%   \state{Beijing}
% \email{junyangwang@bjtu.edu.cn}

% \author{Jitao Sang}
% \affiliation{%
%   \institution{Beijing Jiaotong University}
%   \state{Beijing}
% \email{jtsang@bjtu.edu.cn}

% \author{Yi Zhang, Junyang Wang, Jitao Sang}
% \affiliation{%
%   \institution{School of Computer and Information Technology \& Beijing Key Lab of Traffic Data Analysis and Mining, Beijing Jiaotong University, Beijing, China}
% }

% \email{yi.zhang@bjtu.edu.cn, 21120406@bjtu.edu.cn, jtsang@bjtu.edu.cn}

% \renewcommand{\shortauthors}{Yi Zhang and Jitao Sang}

\newcommand{\myparagraph}[1]{\vspace{3.0pt}\noindent{\bf #1}}
\newcommand{\di}[1]{\textcolor[rgb]{0,0,1}{\textbf{Di}: #1}}
\newcommand{\TODO}[1]{\textcolor[rgb]{0,0,1}{\textbf{TODO}: #1}}
\newcommand{\andreas}[1]{\textcolor[rgb]{0,0,1}{\textbf{Andreas}: #1}}
\newcommand{\shanshan}[1]{\textcolor[rgb]{1,0,0}{\textbf{Shanshan}: #1}}
\newcommand{\bernt}[1]{\textcolor[rgb]{0.82, 0.1, 0.26}{\textbf{Bernt: #1}}}

\makeatletter
\DeclareRobustCommand\onedot{\futurelet\@let@token\@onedot}
\def\@onedot{\ifx\@let@token.\else.\null\fi\xspace}

\def\eg{\emph{e.g}\onedot} \def\Eg{\emph{E.g}\onedot}
\def\ie{\emph{\emph{i.e}}\onedot} \def\Ie{\emph{\emph{i.e}}\onedot}
\def\cf{\emph{c.f}\onedot} \def\Cf{\emph{C.f}\onedot}
\def\etc{\emph{etc}\onedot} \def\vs{\emph{vs}\onedot}
\def\wrt{w.r.t\onedot} \def\dof{d.o.f\onedot}
\def\etal{\emph{et al}\onedot}

\renewcommand*{\@fnsymbol}[1]{\ensuremath{\ifcase#1\or \dagger\or *\or \ddagger\or
   \mathsection\or \mathparagraph\or \|\or **\or \dagger\dagger
   \or \ddagger\ddagger \else\@ctrerr\fi}}


\begin{abstract}
% fairness重要性
Machine learning models often learn to make predictions that rely on sensitive social attributes like gender and race, which poses significant fairness risks, especially in societal applications, such as hiring, banking, and criminal justice. Existing work tackles this issue by minimizing the employed information about social attributes in models for debiasing. However, the high correlation between target task and these social attributes makes learning on the target task incompatible with debiasing. Given that model bias arises due to the learning of bias features (\emph{i.e}., gender) that help target task optimization, we explore the following research question: \emph{Can we leverage shortcut features to replace the role of bias feature in target task optimization for debiasing?} To this end, we propose \emph{Shortcut Debiasing}, to first transfer the target task's learning of bias attributes from bias features to shortcut features, and then employ causal intervention to eliminate shortcut features during inference. The key idea of \emph{Shortcut Debiasing} is to design controllable shortcut features to on one hand replace bias features in contributing to the target task during the training stage, and on the other hand be easily removed by intervention during the inference stage. This guarantees the learning of the target task does not hinder the elimination of bias features. We apply \emph{Shortcut Debiasing} to several benchmark datasets, and achieve significant improvements over the state-of-the-art debiasing methods in both accuracy and fairness.
\end{abstract}

%%
%% The code below is generated by the tool at http://dl.acm.org/ccs.{cf}m.
%% Please copy and paste the code instead of the sample below.
%%




\maketitle

\section{Introduction}
\begin{figure}[t] 
    \centering
    \includegraphics[width=0.45\textwidth]{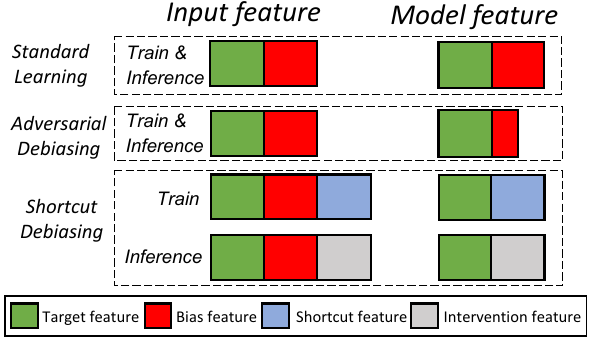}
    \caption{Illustration of the algorithmic bias problem, the conventional debiasing (Adversarial Debiasing), and the proposed Shortcut Debiasing. The removal of bias features in Adversarial Debiasing is potentially hindered by target task learning. Our method avoids this issue by replacing bias features with shortcut features.
    % while our method uses shortcut features to replace bias features in the learning of target task, which avoids the unwitting hindrance of target task learning to debiasing.
    }
    \label{fig:intro}
    % \vspace{-3pt}
\end{figure}
Machine learning algorithms have made notable progress in recent years, and are increasingly being deployed in sensitive and high-stakes environments to make important and life-changing decisions, such as hiring, criminal justice, and banking.
Nevertheless, there is growing evidence~\cite{wang2019racial,grother2019ongoing} that state-of-the-art models can potentially discriminate based on biased attributes such as gender and race, \emph{e.g.}, the popular COMPAS algorithm for recidivism prediction was found biased against black inmates and prone to make unfair sentencing decisions~\cite{lagioia2022algorithmic}.
These ethical issues have sparked a great deal of research into fair machine learning.

% 最近的研究已经意识到不公平现象是因为标准训练的模型(图一中的第一行)继承了真实世界数据中的偏见,进而使模型在决策中不仅使用目标任务特征还学习了偏见特征(\emph{e.g.},gender features),正如图一中第一行所示.
Recent studies~\cite{locatello2019fairness,creager2019flexibly} have shown the learning of bias features (\emph{e.g.}, gender features) is one of the principal causes of unfairness.
As shown in the first row in Figure~\ref{fig:intro}, models with standard learning inherit unfair/biased patterns in training data, and the learned decision rules thus depend on both target and bias features. 
This realization spawned a lot of methods to prevent models from learning bias features.
% 其中,朝着这个目标最直接的方法【】【】是直接删去训练数据中关于偏见特征的特征(\emph{e.g.},直接删除性别特征),然而,这只适用于结构化数据而不适用于视觉数据因为样本中目标特征和偏见特征是紧密耦合.
The most direct method~\cite{d2017conscientious} removes bias features from the training data (\emph{e.g.}, directly removing all information about gender). However, this only works for structured data, because the target features and bias features are tightly entangled in other types of data, such as images.
More typical bias mitigation methods ~\cite{raff2018gradient,kim2019learning,jung2021fair} employ adversarial-based regularization terms to remove bias features encoded in the model. 
Taking Adversarial Debiasing as an example, the models are trained adversarially to discriminate target attribute labels and fail to discriminate bias attribute labels. The goal of these methods is to remove biased information from model representations and enhance invariance to bias attributes.
% However, there is a limitation that they potentially remove useful target information for debiasing resulting in a loss of accuracy Because the target task information has a strong correlation with biased attribute information in biased training data. Also due to this correlation, the training process of the target task itself will promote the representation of bias attributes, resulting in limited debiasing effect. This game between target task learning and debiasing task leads to accuarcy-fairness paradox (\emph{c.f.}, Figure 1(a)).
However, since target features (\emph{e.g.}, \emph{nurse}) and bias features (\emph{e.g.}, \emph{female}) have strong correlations in real-world training data, the optimization of the target task inevitably prevents debiasing operation from removing bias features.
This incompatibility between the target task and the debiasing leads to limited debiasing effects, that is, the bias features encoded in the model cannot be completely removed (see the second row of Figure~\ref{fig:intro}).
% 对于上述的零和游戏，我们以对抗去偏见方法为例进行了描述，可以看到去偏见效果和目标任务的表现是相悖的，我们在图二中描述上述的
% Consider an intuitive example, in the training set of occupation recognition, most nurses are women and most doctors are men. If the model is discouraged from extracting gender (for debiasing), this will result in the model failing to predict occupations well. 
 % removing bias features may be hindered by learning of target features.

% However, there is a limitation that they potentially remove 目标信息 and 导致了准确率的损伤 due 因为目标特征与偏见特征具有强的相关性in biased training data.同样由于这种相关性,目标任务本身训练的过程中又会促进对偏见属性的表示导致去偏见效果有限.这种目标任务学习和去偏见任务之间的博弈导致accuarcy-fairness paradox (\emph{c.f.}, Figure 1(a)).

% 现有方法通过在训练中引入额外的公平性正则项来移除表征 模型对偏见特征的学习【】【】【】.典型的去偏见方法是对抗去偏见家族其中最典型的是和训练之外是否公平性策略介入训练阶段,第一类是在预测中使用直接修改模型输出阻止模型对偏见特征的学习
% To消除视觉识别中的偏见特征, many previous works (Wang et al. 2019b; Kim et al. 2019; Zhang, Lemoine, and Mitchell 2018)为目标任务的学习之外,以正则化项的形式 train models not to discriminate bias attribute labels.
% These works remove information related to protected attributes in data representation, thereby outputting invariant results in terms of the attributes. 

% 第三段,说目的 以及 idea
% In this paper,我们目的是:
% \IEEEpubidadjcol

% 进一步消除偏见特征的关键是：

% To address the incompatibility between fairness and accuracy in debiasing, this work is thus devoted to designing a debiasing schema that does not need to remove biased information (but removes bias features).
% The problem thus transfers to how to remove bias features without compromising the target task's learning of biased information.
The problem thus transfers to how to avoid the incompatibility between the target task and debiasing to further improve the debiasing effect. The incompatibility arises because the learning of information associated with bias attributes is conducive to the optimization of the target task. Therefore, a key prerequisite for addressing this issue is meeting the target task’s learning requirement on biased information.
% Recalling that model bias arises because the learning of features in regard to bias features helps target task optimization, we explore the following research question: Can we leverage shortcut features to replace the role of bias feature in target task optimization for debiasing?  It is important to note that model bias arises because learning features in relation to biased information, such as bias attributes, aids in the optimization of the target task. Therefore, a key prerequisite for addressing this issue is reducing the reliance on biased information when learning the target task.”
% 实现该目的的先验条件是满足目标任务足够去学习 目标任务的学习
Our attention has been drawn to a flaw in machine learning known as \emph{Shortcut}~\cite{geirhos2020shortcut}, where models rely on easy-to-learn accidental features present in the training set but absent in the test set, while neglecting other features.
Motivated by this, we propose \emph{Shortcut Debiasing} that eliminates model bias without being hindered by target task learning.
As illustrated in Figure~\ref{fig:intro}, the key idea of \emph{Shortcut Debiasing} is to construct \emph{shortcut} solutions that employ artificial shortcut features to replace the role of bias features in target task optimization. 
% 通过在满足目标任务对偏见信息需求情况下去偏见，
% By removing bias features without removing biased information, this breaks the incompatibility between the target task and the debiasing task.
By eliminating bias features while fulfilling the target task’s need for biased information, the incompatibility between target tasks and debiasing is broken.
% , so that we does not need to impose additional fairness constraints on bias features to unwittingly harm target features.
% The basic premise that the shortcut feature can act as a shortcut for the biased information is that the shortcut features can provide consistent information with bias features. For this purpose, we attach corresponding shortcut features to samples with different bias attributes to satisfy biased information consistency, \emph{e.g.}, \emph{male} sample and \emph{female} sample are respectively attached with two different shortcut features in training, as shown in Figure~\ref{fig:datademo}.
Benefiting from the fact that the artificial shortcut features are manually controllable, during inference, we can replace shortcut features of samples with \emph{intervention feature} based on causal intervention to eliminate the unfairness caused by shortcut features. Note that we do not need any prior information on test samples, \emph{i.e}., the intervention features imposed on all samples are the same.
The collaboration of shortcut features and intervention features corresponding to training and testing constitutes the \emph{Shortcut Debiasing} scheme.
% The synergy of shortcut features and intervention features allows us to use shortcut features instead of bias features, and further eliminate biased information in shortcut features.

% shortcut效果of代理特征是debiasing的核心，我们使用了XXX
The shortcut effect of shortcut features directly determines the debiasing performance. We conduct shortcut effect analysis and reach an observation: the models do not naturally learn all biased information from shortcut features and still learn biased information from bias features, resulting in the inability to eliminate bias completely. 
% 我们
To address this issue, we propose to maximize the contribution of biased information in shortcut features to the target task to enhance the shortcut effect of shortcut features on biased information in target task learning, which we call \emph{Active Shortcut Debiasing}. 
For measuring the contribution of shortcut features, we borrow the idea of counterfactual explanations~\cite{mothilal2020explaining}. 
% Our Shortcut Debiasing improves over previous methods by avoiding the conflict between debiasing and target task learning in previous methods on debaising performace. 
% 之后我们构造了额外观察发现,一对一不够,模型仍然会从偏见里学习,这是符合我们直觉的
% 因此我们又XXX提升模型对代理特征的依赖,以此模型在学习中会依赖代理
% 最后为了在测试阶段消除代理特征对模型决策的影响,我们基于因果干预将代理特征替换为了干预特征.note that 所有样本中的干预特征都是一致的,是偏见属性无关的和目标属性无关的.

Further, we also provide theoretical justifications on \emph{Active Shortcut Debiasing} from the Bayesian perspective, verifying why shortcut features have the ability to substitute bias features in target tasks.
% 具体的，我们表明在不公平数据上学习到的偏见模型相比于公平数据上学习到的公平模型额外学习了p（y|b），也就是b到y的相关性。
% 从这个视角，我们通过直接使用朴素的代理特征（ie直接attact简易特征）可以被动地代替一部分p（y|b），实现消除偏见。
% 我们进一步提出地active去偏见，实际上通过最大化代理特征对y的贡献，将p(y|b)的注意力完全集中于代理特征。因此，目标任务在不公平数据集上进行训练时不会再使用偏见特征，因为模型对偏见信息学习的容量已经被代理特征所填满，which导致了模型在测试时对偏见特征不变性因此实现了偏见的消除。
% 具体的，我们展示模型学习了数据中的偏见信息
Our justifications demonstrate that, due to the Shortcut Effect Enhancement in \emph{Active Shortcut Debiasing}, the model learns biased information in the training set only from shortcut features rather than from both shortcut and bias features. This guarantees that the intervention on the shortcut features during the inference stage removes all bias from the model.

% 在使用代理特征代理消除偏见的过程中,整个方法工作像是一个普通的学习过程,我们没有使用额外的操作对模型表征进行公平性约束,因此我们叫做代理学习

We summarize our main contributions as follows:
% \vspace{-5pt}
\begin{itemize}
        \item We propose a novel debiasing method \emph{Shortcut Debiasing}, which constructs \emph{shortcut} solutions to employ shortcut features to replace bias features in the target task's learning. This avoids the competition between target task learning and debiasing in previous methods.
    % \vspace{-3pt}
    % 我们提出了shortcut learning 方法.这个方法构建了能够提供给模型偏见信息的代理特征,并主动加强代理特征对目标任务贡献来提升代理特征对偏见特征的代理作用.结合测试阶段的因果干预,我们消除了代理特征的包含的偏见信息带来的影响.
    % \item We introduce shortcut effect enhancement that actively enhances the contribution of shortcut features to the target task to improve the shortcut effect of shortcut features on bias features.
    % 尽我们所知，我们是首个开发了深度学习的shortcut这一劣性性质的良性应用。对去偏见问题的理论分析确保了shortcut应用于debiasing的可行性。

    %理论保证我们实现shortcut，干预特征保证了消除
        % \vspace{-3pt}

    \item We are the first to explore the benign application of the shortcut properties of machine learning. Theoretical justifications ensure shortcut features can replace bias features, and causal intervention eliminates the bias brought by shortcut features.

    \item Extensive experiments demonstrate that our method significantly outperforms baselines on both accuracy and fairness. 
    % The effectiveness of multiple bias attributes debiasing is also verified.
    
\end{itemize}

% \section{Background and Related Work}
\section{Related work and Background}
\subsection{Bias Mitigation} 
% AI模型中的Gender 和 racial偏见已经被documented.
% Gender and race bias in AI models have been documented~\cite{yao2017beyond,zhao2017men}. 
% 以Equalodds(详细描述在sec4)为代表的偏见衡量方法揭示了深度学习exhibit social bias towards certain demographic groups.
% Bias measurement approach such as Equalodds reveal that AI models exhibit social bias towards certain demographic groups.
% 有很多工作已经被提出to tackle fairness problem in machine learning.
% Much research work has been done on mitigating model bias.
% 现存的公平性方法主要可以被分为三类根据消除方式.
Existing methods for mitigating bias can be broadly categorized into three groups based on the stage of the training pipeline to which they are applied. Pre-processing methods~\cite{louizos2015variational,quadrianto2019discovering} refine dataset to mitigate the source of unfairness before training; in-processing methods~\cite{elkan2001foundations,jiang2020identifying} introduce fairness constraints into the training process; and post-processing methods~\cite{kamiran2012decision,pleiss2017fairness} adjust the prediction of models according to fairness criterion after training. Among them, this paper focuses on in-processing methods.

% 大多数/典型 中处理 做法
Typical in-processing bias mitigation studies employ additional fairness constraints as regularization terms for mitigating bias.
~\cite{zhang2018mitigating,wang2019balanced} enforce the model to produce fair outputs with adversarial training techniques by minimizing the ability of a discriminator to predict the bias attribute.
~\cite{kim2019learning} further minimizes the mutual information between representation and bias attributes to eliminate their correlations for debiasing. ~\cite{tartaglione2021end} devises a regularization term with a triplet loss formulation to minimize the entanglement of bias features. ~\cite{jung2021fair} tries to distill fair knowledge by enforcing the representation of the student model to get close to that of the teacher model averaged over the bias attributes.
% 存在的问题 然而,强相关性导致去偏见效果和
% However, the high correlation between target task and bias attributes that exist in the data itself leads to the limited accuracy in debiasing.
Meanwhile, some methods try to transform the target task into not actively extract biased information. ~\cite{wang2020towards} trains different target models separately for each group in terms of bias attributes so that the target task does not attempt to rely on bias features. ~\cite{du2021fairness} trains the classification head with neutralized representations, which discourages the classification head from capturing the undesirable correlation between target and biased information. ~\cite{park2022fair} proposes to employ contrastive learning to learn bias-invariant representations for downstream tasks.
% 然而这些方法仍然难以解决悖论,因为偏见存在数据本身
% However, these methods still suffer from fairness-accuracy paradox due to the inherent bias of the training set.
% 在本文中,我们解决数据本身偏见通过构建偏见特征的代理特征.
% In this paper, our method leverages shortcut features to shortcut the biased information in data for unraveling the fairness-accuracy paradox.
% However, the high correlation between target task and bias attributes that exist in the data itself leads to the limited debiasing performance.
However, the high correlation between target task and bias attributes leads to bias mitigation incompatible with target task learning.

\subsection{Shortcut Learning} 
The \emph{shortcut} property of machine learning~\cite{geirhos2020shortcut}, one of the defects of modern machine learning, refers to the fact that a machine learning model may only use accidental easy-to-learn features that exist in the training set but not in the test set during training, and the machine learning model performs extremely poorly on the test set. 
The \emph{shortcut} learning property of machine learning exists not only in computer vision problems~\cite{wen2022fighting,robinson2021can} but also in natural language processing problems~\cite{du2021towards}.

Due to the critical risk to machine learning algorithms, \emph{shortcut} has been viewed as malignant in default, which naturally gives rise to considerable attention on circumventing shortcut solutions. 
% The underlying mechanism of shortcut learning property may be because the model tends to learn features of easy patterns to shortcut complex patterns in the data~\cite{arpit2017closer}.
Much research work~\cite{saranrittichai2022overcoming,minderer2020automatic} has begun to try to eliminate shortcuts to achieve more robust models. Yet, apart from continuously proposing new ways to prevent model shortcuts, no work has explored exploiting shortcuts in other-than-malignant applications.
This paper provides an alternative perspective to consider \emph{shortcut} and explore whether we can exploit it in benign applications, \emph{i.e}., fairness.
% Towards this goals, we propose the mode that actively replace harmful features with \emph{shortcut feature} and eliminate shortcut features at inference time.

\subsection{Causal Intervention} 
% \myparagraph{Causal intervention.} 
% 因果推断
Causal intervention~\cite{pearl2018book} seeks the true causal effect of one variable on another~\cite{keele2015statistics}. It not only serves as a means of interpreting data but also offers solutions for achieving desired objectives through the pursuit of causal effects.
Causal intervention has been widely used across many tasks to improve the reliability of machine learning models. Backdoor adjustment~\cite{pearl2014interpretation} is one of the most widely used implementations of causal intervention. The technical principle of Backdoor adjustment will be expanded upon in detail in Section 3. Next, let's review the application scenarios of backdoor adjustment.

% Taking Figure~\ref{fig:causalgraph}(c) as an example, different nodes represent different variables. Given target feature $X_T$ and other features $P$, we hope the model’s output $T$ being faithful only to the target feature $X_T$.

Several studies~\cite{yue2020interventional,zhang2020causal} leverage backdoor adjustment for eliminating the confounding factor in few-shot classification and weakly supervised segmentation. ~\cite{wang2020visual} employs backdoor adjustment to train feature extractors with commonsense knowledge by eliminating unreliable knowledge.
~\cite{tang2020long} uses backdoor adjustments to eliminate the long tail caused by the harmful direction included in the SGD momentum in the optimizer. ~\cite{qi2020two} obtains more core visual dialogue clues based on backdoor adjustments during training.
% 【】通过利用后门调整干预来训练具有常识知识的特征提取器,【】【】利用后门调整来训练鲁棒的小样本分类和弱监督分割.
% 与已有工作通过干预启发的训练不同,我们将干预应用于测试阶段来消除较好继承了训练数据的模型中的偏见.通过测试阶段的干预策略,我们可以避免在训练中消除偏见带来的目标任务和去偏见任务的零和博弈.
% Different from existing work to inspire training strategies with causal interventions, 
% we leverage causal intervention combined with the shortcut feature strategy into the model inference stage to eliminate model bias.
Unlike existing work that applies causal intervention in the training phase, we apply causal intervention in the inference phase to remove the influence of shortcut features.

\begin{figure}[t] 
    \centering
    \includegraphics[width=0.46\textwidth]{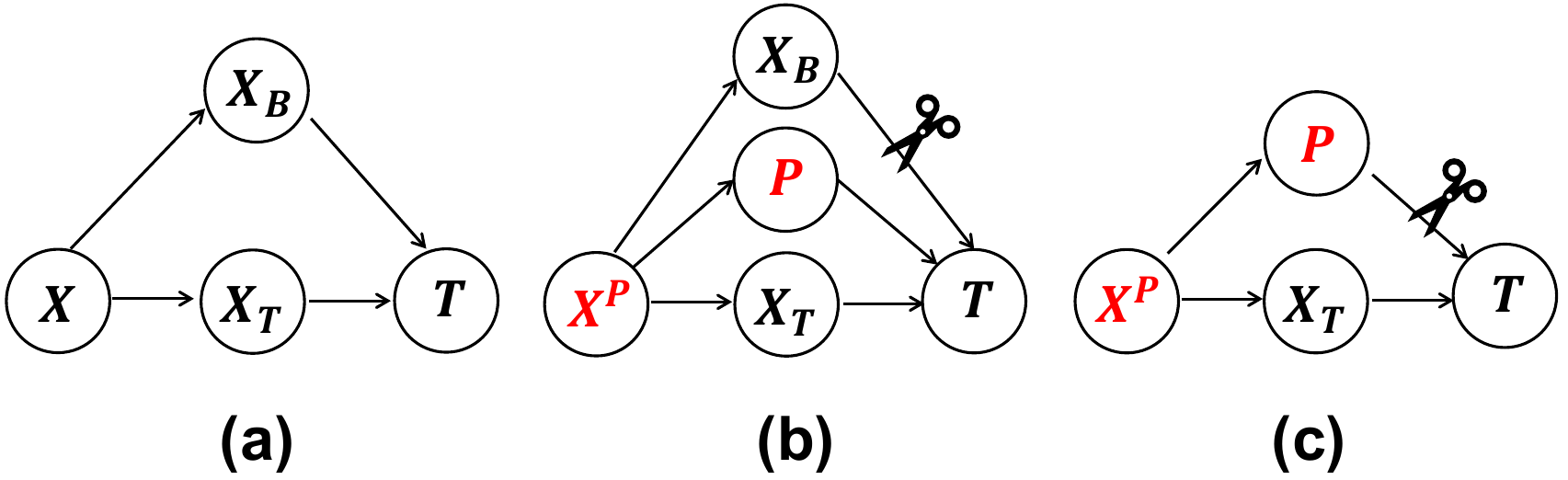}
    \caption{\textbf{The causal graph of the proposed model.} (a) The output $T$ of biased model is directly affected by the target feature $X_T$ and bias feature $X_B$ in input $X$. The training and inference stage of \emph{Shortcut Debiasing} are illustrated in (b) and (c), respectively.
    }
    \label{fig:causalgraph}
    % \vspace{-10pt}
\end{figure}

\begin{figure}[t] 
    \centering
    \includegraphics[width=0.43\textwidth]{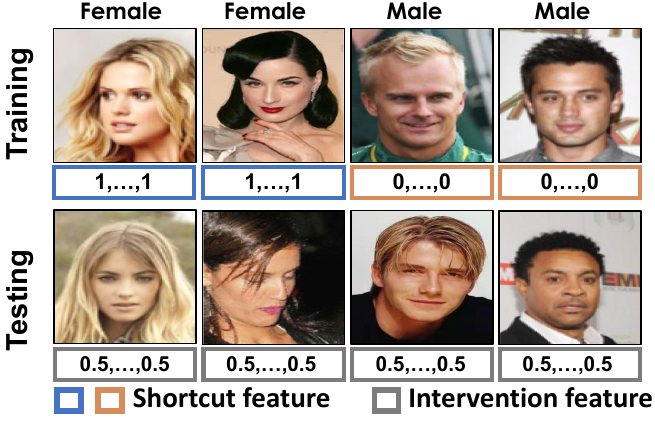}
    \caption{\textbf{Examples of Naive Shortcut Debiasing for eliminating gender bias.} In training, images of different genders are assigned different shortcut features, while all images are assigned the same intervention feature in testing.
    }
    \label{fig:datademo}
    % \vspace{-10pt}
\end{figure}

\begin{figure*}[t]
    \centering
    \includegraphics[width=0.98\linewidth]{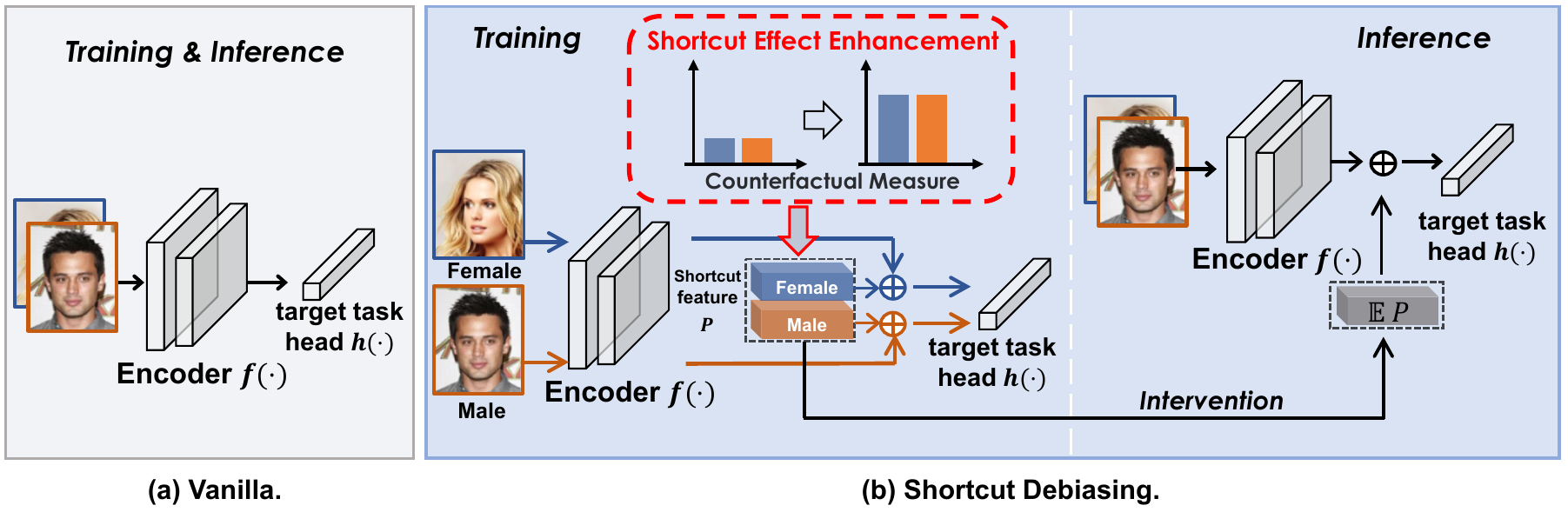}
    % \vspace{-10pt}
    \caption{\textbf{Illustration of Vanilla and Shortcut Debiasing methods.} Active Shortcut Debiasing contains additional shortcut effect enhancement module (highlighted with \textcolor{red}{red} dash line) than Naive Shortcut Debiasing.
    }
    \label{fig:overview}
    % \vspace{-5pt}
\end{figure*}

\section{METHODOLOGY}
\subsection{Overview}
% 我们关注视觉识别任务中的公平性
This paper focuses on fairness in visual recognition. A fair visual recognition model is one that makes predictions based solely on the feature $X_T$ of the input $X$. However, when trained on biased data, standard learning models may learn to rely on bias features $X_B$ of the input $X$, as illustrated in Figure~\ref{fig:causalgraph}(a). The aim of fairness is to eliminate the model’s dependence on bias features $X_B$, ensuring that the output $T$ is independent of these features. To achieve this goal, in the fairness research problem setting, both the target task label $t \in T$ and the bias attribute label $b \in B$ of the input image $x \in X$ are provided in the dataset.

% We explore the benign application of shortcut properties to debiasing scenarios in the form of human-specified shortcut solutions 有偏见决策规则.

% The shortcut solutions of machine learning have been recognized as one of the flaws of machine learning, because machine learning may only learn accidental features that are easy to learn in the training set, which will lead to random guesses in the test set without these accidental features.
% Inspired by this, we explore the benign application of shortcut properties to debiasing scenarios in the form of human-specified shortcut solutions.
Although the shortcut property is often viewed as a disadvantage in machine learning, we explore its potential advantages in addressing fairness issues by constructing artificial shortcut solutions for the bias decision paths $X_B \rightarrow T$. Specifically, we propose \textbf{Shortcut Debiasing} that performs debiasing in two stages: (1) During the training of the target task, the model is guided to preferentially use shortcut features $P$ with easy-to-learn patterns to learn biased information. As a result, the model no longer needs to pay attention to bias features $X_B$, and the path $P \rightarrow T$ becomes a shortcut for path $X_B \rightarrow T$ (see Figure~\ref{fig:causalgraph}(b)); (2) During testing, the causal intervention mechanism is introduced to eliminate the influence of shortcut features $P$ on output $T$ (see Figure~\ref{fig:causalgraph}(c)).

The direct way to realize \emph{Shortcut Debiasing} is to attach pre-defined shortcut features with easy-to-learn patterns to original features, and then the model learns from this composite feature (see Figure~\ref{fig:overview}). This leads to the basic version of our solution, which we call \emph{Naive Shortcut Debiasing} and will be introduced in the next subsection. 
% 然而,之后的分析表明模型并不会自然的优先学习代理特征.
However, the subsequent empirical analysis shows that shortcut features with easy-to-learn patterns do not naturally replace bias features.
% 去提升代理特征的优先级
In order to ensure the shortcut effect of shortcut features, we further propose to enhance the target task's attention to shortcut features, which we call \emph{Active Shortcut Debiasing} as a complete version of our solution.

% 最后，我们理论证明了
Then, we also provide theoretical justifications for \emph{Active Shortcut Debiasing} from a Bayesian perspective, verifying that shortcut features have the ability to fully substitute bias features in the learning of target tasks.

\subsection{Naive Shortcut Debiasing}
\myparagraph{Training with shortcut features.}\hspace{1.5mm}
% In fair visual recognition problem, input $x \in X$ is given two types of labels: target task attribute $t \in T$ and bias attribute $b \in B$. 
% Under the motivation that use shortcut features to replace 模型对偏见特征的依赖,我们需要构造与偏见特征分布一致的代理特征
% Under the motivation that to use shortcut features to replace bias features, t
The premise of realizing a shortcut to bias features $X_B$ is that the shortcut features $P$ should provide the same biased information as the bias features.
To this end, we construct shortcut features that are consistent with the distribution of bias features (\emph{e.g., gender}), as depicted in Figure~\ref{fig:datademo} (\emph{Training}).
Specifically, considering that simple patterns are easier to learn, we preset the shortcut features $p_b \in P$ as all zeros and all ones vectors. Then, We utilize bias attribute label $b$ to select the corresponding shortcut feature $p_b \in P$ to ensure information consistency between the same sample's shortcut features and bias features.

% 训练中
During training, we concatenate the image feature $f(x)$ encoded by the encoder $f(\cdot)$ and the corresponding shortcut features $p_b$ to obtain the composite feature $\{{f(x)},p_b\}$, as illustrated in Figure~\ref{fig:overview}(b).
% we contact the corresponding shortcut feature $p_b$ to the feature representation of the sample $x$.we train the target task in this composite data: 
% Note that we do not put the shortcut feature as a patch in the image because it will occlude part of the image information.
Then, we train the target task using the composite feature:
\begin{equation}\label{targetlearn}
\min \limits_{f,h}  L_{\text{target}}(h(\{{f(x)},p_b\}), t), 
\end{equation}
where $L_{\text {target}}$ is the target task loss (\emph{i.e}., cross-entropy loss), and $t$ is the target task label of $x$. 

\myparagraph{Inference with intervention feature.}\hspace{1.5mm} 
By learning shortcut features during training, the model’s reliance on bias information shifts from the bias features $X_B$ to the shortcut features $P$. However, this introduces a new issue: as illustrated in Figure~\ref{fig:causalgraph}(c), shortcut features $P$ become a new source of model bias in inference:
% By training with shortcut features, the dependence of trained models on biased information is based on shortcut features $P$ rather than bias features $X_B$.
% However, as shown in Figure~\ref{fig:causalgraph}, this introduces $P$ as a new source of biased information in model inference:
\begin{equation}\label{equbias}
Pr(T \mid X)=Pr(T \mid X_T, P)
\end{equation}
% of different shortcut features that correspond to different bias labels, such as male and female.

% A unbiased model should only be predict based on $X_T$ in image X, not on any biased information about $B$ including $P$.
And now, we need to remove the model bias brought about by shortcut features $P$. 
By taking advantage of the controlled nature of the shortcut features, we are able to eliminate their contribution to the model output $T$ through the causal intervention.
% 受益于人工代理特征是受控的，这使得基于因果干预来消除人工代理特征对模型输出的贡献成为了可能。

% 因果干预被设计用于消除混杂因子对建模两个变量关系的影响。在本文中，要建模关系的两个变量是xt和t，混杂因子为p。直接的干预方案是人工修改xt（被称作 do 操作）观察t的变化来建模xt与t的关系：

Causal intervention is designed to remove the influence of confounding factors on modeling the relationship between two variables. In this paper, the two variables to model the relationship are $X_T$ and $T$, and the confounding factor of $P$. The direct causal intervention is to manually modify (called \emph{do} operation) $X_T$ to observe the probability relationship between $do(X_T)$ and $T$: 
\begin{equation}
Pr(T \mid do(X_T))
\end{equation}

% 然而在推理中xt作为真实特征是无法被真实编辑的。后门调整提供了另一条间接干预的思路，通过平均地观察在不同p下的模型输出来如果是在物理世界无法直接Do操作指的是如果模型的输入为人工设置为xt,而不存在即为Do操作，do操作的目的是
Nonetheless, given the extensive range of potential values for $X_T$, it is impractical to intervene directly on every different $X_T$. The \emph{backdoor adjustment} provides a solution for indirect intervention by averaging the contribution of different $P$ to the model output $T$, preventing the introduction of causal effects of $P$ to $T$. The causal effects of the target feature $X_T$ to model output $T$ can be derived:
\begin{equation}
Pr(T \mid do(X_T))=\sum_{b}^{} Pr(T \mid X_T, p_b) Pr(p_b),
\end{equation}
where $p_b$ is the shortcut feature corresponding to bias labels $b$. The underlying mechanism of backdoor adjustment is to force $X_T$ to incorporate every $p_b$ fairly, to neutralize the impact of a specific $p_b$. 

For low computation cost during inference, we replace $\sum_{b}Pr(T\mid X_T, p_b) Pr(p_b)$ with $Pr(T\mid X_T,\mathbb{E}_{\boldsymbol{b}} \left[p_b\right])$ due to \emph{NWGM} linear approximation proved in~\cite{xu2015show}:
\begin{equation}\label{inter}
Pr(T \mid do(X_T))=Pr(T \mid X_T, \mathbb{E}_{\boldsymbol{b}} \left[p_b\right])),
% \mathbb{E}_{\boldsymbol{z}}\left[g_{y}(\boldsymbol{z})
\end{equation}
where $\mathbb{E}_{\boldsymbol{b}} \left[p_b\right]$ denotes  mathematical expectation of $p_b$ with respect to $b$, \emph{i.e}., the mean of $p_{\boldsymbol{male}}$ and $p_{\boldsymbol{female}}$. This mean vector is referred to as the \emph{intervention feature}.

In the implementation of the algorithm, the intervention feature $\mathbb{E}_{\boldsymbol{b}} \left[p_b\right]$ is directly utilized in place of the shortcut features $p_b$ associated with the test sample during the inference stage. As shown in Figure~\ref{fig:datademo} (\emph{Testing}), we utilize a vector comprised entirely of 0.5. It is important to note that utilizing intervention features presents an additional advantage: there is no requirement to ascertain the bias label of samples during inference.

\subsection{Analysis of the Shortcut Effect}
% We examine the debiaisng performance of Naive Shortcut Debiasing.
Taking face attribute recognition in CelebA~\cite{liu2015deep} as an example of target task, we examine the gender debiasing performance of \emph{Naive Shortcut Debiasing} on \texttt{Blonde} and \texttt{Attractive} attribute recognition tasks. 
Accuracy and model bias (gender bias) are reported in Table~\ref{analysis}. Section~\ref{Experiments} provides a detailed description of the model bias measurement.

For the \texttt{Blonde} recognition task, \emph{Naive Shortcut Debiasing} significantly improves both fairness (\emph{i.e}., lower model bias) and accuracy compared to Vanilla. 
However, neither model bias nor accuracy changes significantly in \texttt{Attractive} recognition task.
Given that the same configuration is used for both tasks, we conjecture that the reason for the performance inconsistency of \emph{Naive Shortcut Debiasing} on the two tasks might lie in the differences in the learning of shortcut features. 
To test this conjecture, we look into the effect of shortcut features on model output through counterfactual comparison, \emph{i.e}., how the change of shortcut features in test samples influences the model’s decision-making:
% \begin{equation}\label{counterp}
% \begin{aligned}
% Counter&@P =    \\ 
% & \mathbb{E}_i \left| Pr(Y\mid x_i, p_{\boldsymbol{male}})-Pr(Y\mid x_i, p_{\boldsymbol{female}})\right|,
% \end{aligned}
% \end{equation}
\begin{equation}\label{counterp}
\begin{aligned}
Co&unter@P \\
=& \mathbb{E}_i \left| Pr(T=t_i\mid x_i, p_{\boldsymbol{male}})-Pr(T=t_i\mid x_i, p_{\boldsymbol{female}})\right|,
\end{aligned}
\end{equation}
where $p_{\boldsymbol{male/female}}$ is shortcut feature regarding male or female, $t_i \in T$ is the target label of sample $x_i$, $Pr(T\mid x_i, p_{\boldsymbol{male/female}})$ is the model output.

% 解释结果 我们观察到去偏见效果好的blonde识别有着,继而模型不会将对偏见属性的依赖转移到代理特征上.

Table~\ref{analysis} shows that the model with Naive Shortcut Debiasing  has no significant dependence on shortcut features (lower $Counter@P$) for poor debiasing performance task (\texttt{Attractive}). In contrast, the task with high debiasing performance (\texttt{Blonde}) corresponds to high $Counter@P$. These findings suggest that models do not always prioritize learning simple pattern shortcut features $P$, and may even prioritize learning bias features $X_B$. As a result, shortcut features $P$ cannot fully replace bias features $X_B$.

% As reported in Table~\ref{analysis}, we note that in Naive Shortcut Debaisng, for poor debiasing performance task (Attractive), the model has no significant dependence on shortcut features(lower $Counter@P$). However, high debiasing performance task (Blonde) corresponds to high $Counter@P$. 
% This depicts that models do not always learn the simple pattern shortcut features preferentially, even prioritizing learning of bias features, and therefore the shortcut features $P$ cannot replace the model's reliance on bias features $X_B$.

\begin{table}[t]

   \caption{The accuracy (in $\%$), model bias (in $\%$, described in Sec.4) and Counter@$P$ of Naive Shortcut Debiasing (\textit{Naive SD}).}
       % \vspace{-5pt}

%   For each bias, we used four text templates to measure bias and calculate the aug.For each bias, we calculated using four text templates: Sentence w/o specify(Vanilla), Sentence w/ specify(Naive Proxy Debiasing), Phrase w/o specify(P$_\text{W/O}$) and Phrase w/ specify(P$_\text{W/}$)
  
%     \small
% 	\centering	
%     \aboverulesep = 0.46mm
%     \belowrulesep = 0.46mm
% \setlength\tabcolsep{5pt}
	\resizebox{0.48\textwidth}{!}
{
	\begin{tabular}	{l | l |  c  c  c  }
		\toprule
% 		\multirow{2}{*}{\multirow{2}{*}{VLPs}}
% 	  &  \multicolumn{3}{c}{TR}& \multicolumn{3}{c}{IR} \\
% 	 \midrule
	 \textbf{Task}&\textbf{Method}&\textbf{Acc.($\uparrow$)}&\textbf{ Bias($\downarrow$)}&\textbf{Counter@P}\\
    \midrule
	\multirow{2}{*}{\texttt{Blonde}}  &\textit{Vanilla}& 78.77 &40.82&- \\
	 &\textit{Naive SD}& \textbf{91.53}\footnotesize{(+12.76)} &\textbf{10.09}\textcolor{red}{\footnotesize{(-30.73)}}&\textcolor{red}{0.26} \\
	     \midrule
	\multirow{2}{*}{\texttt{Attractive}}  &\textit{Vanilla}& 76.72&26.24&- \\
	&\textit{Naive SD} & \textbf{77.25}\footnotesize{(+0.53)} &\textbf{24.06}\textcolor{red}{\footnotesize{(-2.18)}}&\textcolor{red}{0.01} \\ 
		\bottomrule
	\end{tabular}
 	}
 	
       % \vspace{-10pt}

% 	\caption
% 	{
% % 	\small	
% 		
% 	}
	\label{analysis}
\end{table}

\subsection{Active Shortcut Debiasing}
\myparagraph{Training with shortcut features.}\hspace{1.5mm}
The empirical analysis above suggests that shortcut features with simple patterns may be trivial to target task. To address this issue, we propose the \textbf{\emph{Shortcut Effect Enhancement}} module (red dash line in Figure~\ref{fig:overview}), which guides the model to actively learn biased information from shortcut features during target task training (Eqn.~\ref{targetlearn}). This way, the target task does not need to focus on bias features $X_b$.

% 代理特征的优先学习是。。。XXX

% To this end, we to \emph{Naive Shortcut Debiasing}. 
% apply gradcam
% As illustrated in Figure 2, our model includes three trainable modules: shortcut features, intervention features and classifier.

Instead of using preset shortcut features $P$ and allowing the target task head (\emph{i.e}., $f(\cdot)$) to learn them passively, we set the shortcut features $P$ to be trainable and guide the target task’s reliance on bias information from bias features to the shortcut features.

% Specifically, instead of using preset shortcut features and target task heads (\emph{i.e}., $f(\cdot)$) that passively learn shortcut features, we employ trainable shortcut features and actively train target task heads to better satisfy the target model's reliance on biased information in shortcut features.

To accurately measure the model’s dependence on shortcut features and guide its reliance on them, we borrow and revise the feature attribution strategy based on counterfactual analysis~\cite{lang2021explaining,zhangcounter}, which measures the importance of shortcut features by counterfactually changing them:
\begin{equation}\label{counter}
\alpha^{c}=Y_c(x, p_{\tiny{b}})-Y_c(x, anchor),
\end{equation}
where $\alpha^{c}$ represents the importance of shortcut features to target class $c$ (\emph{e.g.}, attractive or non-attractive) from the model’s perspective, $Y_c(\cdot)$ is the logit output corresponding to class $c$, $p_b$ is the trainable shortcut features associated with the bias label $b$ of $x$. The $anchor$ is the preset counterfactual contrast point, obtained through random initialization, which serves as the counterfactual feature for $p_b$.

% we use trainable shortcut features for better meet the model on shortcut features. 
% 我们应该XXX 然后其他

To enhance the importance of shortcut features on the target task, we propose Shortcut Effect Enhancement. This enhancement is iteratively performed in parallel with the target task training:
% Specifically, the following two steps iterate during the training process:
\begin{itemize}
\item Target task training: 
$\min \limits_{f,h}  L_{\text{target}}$ following Eqn.~\eqref{targetlearn}

\item Shortcut effect enhancement: 
\begin{equation}\label{active}
\max \limits_{P,h}  \frac{\exp \left(\alpha^{t}\right)}{\sum_{i=1}^{T} \exp \left(\alpha^{i}\right)},
\end{equation}
where $t$ is the target label of input $x$. We aim to maximize the importance of shortcut features by optimizing both shortcut features $P$ and target task head $h(\cdot)$ using softmax normalization.
\end{itemize}
% Toward reinforce the importance of shortcut features, we can update the shortcut feature by maximize the contribution of shortcut feature on target label:
% \begin{equation}
% \mathop{max}\limits_{p}\ { {Contri}_y}
% \end{equation}
% where . By iteratively optimizing Eqn.~\ref{targetlearn} and Eqn.~\ref{active}, the trainable shortcut features $P$ are guaranteed to hold the shortcut effect to bias feature $X_b$ continuously.
% 随着模型使用
% Same as Eqn.~\ref{targetlearn}, we composite image $x$ and shortcut feature $p$ as input to train target task.

In this way, the shortcut effect of shortcut features is adaptively reinforced based on the current state of target task head $h(\cdot)$.
% and thus guaranteed to preferentially use the shortcut feature over the bias feature in the next step's target task training. 

\myparagraph{Inference with intervention feature.} \hspace{1.5mm}
Similar to Eqn.~\ref{inter} in \emph{Naive Shortcut Debiasing}, we substitute the shortcut feature with the intervention feature $\mathbb{E}_{\boldsymbol{b}} \left[p_b\right]$. The key difference is that the shortcut features $P$ here are obtained through training. 
% The complete procedure is described in the Appendix.
%, we use the trained $P$ to compute $\mathbb{E}_{\boldsymbol{b}} \left[p_b\right]$.

\subsection{A Bayesian perspective}
% 在这一节，我们提供了一个贝叶斯视角的理解and展示模型会只从代理特征中学习偏见特征而不是同时从代理特征和偏见特征中。这保证了推理时对代理特征的干预消除了模型的全部偏见。
In this section, we provide a Bayesian perspective of \emph{Active Shortcut Debiasing} and show that the model will learn biased information exclusively from shortcut features, rather than from both shortcut features and bias features. This ensures that the intervention on the shortcut features in inference removes all bias from the model.

In visual recognition, we are interested in estimating the conditional probability $Pr(t|x)$.
% , which is generally modeled by Softmax function as a multinomial distribution $\phi$:
% \begin{equation}\label{softmax}
% \phi_j=\frac{e^{\eta_j}}{\sum_{i=1}^k e^{\eta_i}},
% \end{equation}
% where Softmax function maps a model’s class-$j$ output $\eta_i$ to the conditional probability $\phi_j$.
For fairness problems, $Pr(t|x)$ can be interpreted from a Bayesian perspective as:
\begin{equation}\label{bayes}
Pr(t| x)=Pr(t| x, b)=\frac{Pr(x| t, b)}{Pr(x|b)} Pr(t|b),
\end{equation}
where $b$ is an implicit input about biased information to the model, that is, the information provided by shortcut features $P$ and bias features $X_b$ (Figure~\ref{fig:causalgraph}). 

Assuming that all instances in both the fair training set (\emph{i.e}., bias labels are independent of target labels) and unfair training set (\emph{i.e}., bias labels are relative to target labels) are generated from the same process $Pr(x|t,b)$, the Bayesian inference difference between the fair and unfair training set is $Pr(t|b)$ and $Pr(x|b)$. For $Pr(x|b)$, we analyzed that $Pr(x|b)$ was not a cause of model bias.

For $Pr(t|b)$ in Eqn.~\eqref{bayes}, counterfactual analysis (Eqn.~\eqref{counter}) is used to eliminate the effect of $\frac{Pr(x| t, b)}{Pr(x|b)}$ because the counterfactual change of $P$ does not change the generation process of $x$ (\emph{i.e}., $\frac{Pr(x| t, b)}{Pr(x|b)}$). The subtraction of the logit output space is equivalent to the division of the probability output space, and the subsequent Softmax function (Eqn.~\eqref{active}) computes the correlation $Pr^P(t|b)$ between the biased information in shortcut features $P$ and the model output $T$.

\begin{table*}[th]

\caption{The Bias Accuracy (in $\%$, $\uparrow$), Fair Accuracy (in $\%$, $\uparrow$), and model bias (in $\%$, Equalodds, $\downarrow$) of models trained on CelebA. Here T and B respectively represent target and bias attributes.  Here \textit{a}, \textit{bl}, \textit{bn}, \textit{ps}, \textit{m}, and \textit{y} respectively denote \textit{attractive}, \textit{blonde}, \textit{bignose}, \textit{pale skin}, \textit{male}, and \textit{young}.}
% \vspace{-5pt}

\centering
\resizebox{0.99\textwidth}{!}{
\begin{tabular}{c|ccccccccccccccccccc}
\toprule
\multirow{3}{*}{Method } &\multicolumn{3}{c}{T=\textit{a} , B=\textit{m}} && \multicolumn{3}{c}{T=\textit{bl} , B=\textit{m}} &&  \multicolumn{3}{c}{T=\textit{bn} , B=\textit{y}} &&  \multicolumn{3}{c}{T=\textit{ps} , B=\textit{y}} && \multicolumn{3}{c}{\textbf{Avg.}} \\ \cmidrule[0.5pt]{2-4} \cmidrule[0.5pt]{6-8} \cmidrule[0.5pt]{10-12} \cmidrule[0.5pt]{14-16} \cmidrule[0.5pt]{18-20} 
& \multirow{2}{*}{BiasAcc.}  & \multirow{2}{*}{FairAcc.} & \multirow{2}{*}{Bias} && 
\multirow{2}{*}{BiasAcc.}  & \multirow{2}{*}{FairAcc.} & \multirow{2}{*}{Bias} && 
\multirow{2}{*}{BiasAcc.}  & \multirow{2}{*}{FairAcc.} & \multirow{2}{*}{Bias}&& 
\multirow{2}{*}{BiasAcc.}  & \multirow{2}{*}{FairAcc.} & \multirow{2}{*}{Bias}&& 
\multirow{2}{*}{BiasAcc.}  & \multirow{2}{*}{FairAcc.} & \multirow{2}{*}{Bias}
\\
\\ \cmidrule[0.5pt]{1-20} \morecmidrules\cmidrule[0.5pt]{1-20}
\textit{Vanilla} &81.79&  76.72 & 26.24 &&91.60&78.77 & 40.82 &&82.67& 75.39 & 17.78  &&87.96& 88.11 & 14.07&&86.01 & 79.74  & 24.72    \\ \cmidrule[0.5pt]{1-20}
\textit{AdvDebias} &75.45& 77.54 & 11.56 &&87.57&79.24 & 33.44 &&82.48&  71.07 & 7.11  &&\textbf{88.56}& 89.85  & 3.50&&83.51&79.45&13.90\\
\textit{LNL} &\textbf{81.61}&  76.91 & 26.43 &&91.30&79.55 & 33.17 &&\textbf{83.99}& 73.86 & 16.53  &&83.81& 88.00 & 10.70&&85.37& 79.58 & 21.70 \\ 
\textit{EnD} &81.44&  77.11 & 24.64 &&85.41&81.47 & 33.73 &&83.96& 74.18 & 17.65  &&87.50& 89.39 & 10.80&&84.57& 80.53 & 21.70 \\
\textit{MFD} &80.93&  77.22 & 20.17 &&88.50&79.85 & 38.84 &&82.54& 75.39 & 16.12  &&82.23& 87.92 & 14.06&&83.55& 80.09 & 22.29 \\ 
\textit{DI} &80.67&  77.53 & 23.01 &&91.42&91.44 & 7.76 &&82.86& 73.98 & 10.63  &&87.60& 90.19 & 4.53&&85.63&83.26 &11.48    \\
\textit{RNF} &80.10& 75.88 & 24.01 &&86.24& 79.07 & 40.15 &&81.70& 73.19 & 14.36  &&86.19& 86.74 & 14.98&&83.55& 78.72 & 23.37   \\ 
\textit{FSCL} &81.33& 79.79 & 9.86 &&88.96& 86.04 & 7.81 &&83.50& 72.88 & 5.88  &&85.34& 88.97 & \textbf{1.18}&&84.78& 81.92 & 6.18   \\%\cmidrule[0.5pt]{1-18}
\cmidrule[0.5pt]{1-20} 
% \textit{FSCL} & 11.5 & 79.1 && 13.0 & 79.1 && 7.0 & 82.1 && 6.4 & 83.8 && 3.8 & 82.7 && 1.8  & 82.0     \\ 
\textit{Active SD} & 81.43&\textbf{80.54} & \textbf{5.18} && \textbf{91.70} &\textbf{92.02} & \textbf{4.97} &&82.41& \textbf{77.80} & \textbf{5.38}  &&87.92&\textbf{90.58} & 1.23&& \textbf{85.86} & \textbf{85.23}&\textbf{4.19}   \\ \bottomrule
\end{tabular}
}

\label{table:celeba}
\end{table*}

Finally, in Active Shortcut Debiasing, we train the above $Pr^P(t|b)$ to approximate the distribution $Pr(t|b)$ of the training set (Eqn.~\eqref{active}). The resulting $Pr^P(t|b) \approx Pr(t|b)$ means that the model will learn biased information only from shortcut features. 

%在深度学习中，softmax被用于输出属于第几累的概率，比如xxx符号
%在有偏见的训练集下，从贝叶斯推理视角，xxx符号还可以被解释为：

% 其中xxx是重要定义，

\section{Experiments}\label{Experiments}
% In this section, we first introduce fairness protocol, the datasets and baselines.
% We evaluate the effectiveness of our method, followed by extensive debiasing experiments in multiple bias attributes and ablation studies.

\subsection{Experiment Setup}
\myparagraph{Metrics.} Many fairness criteria have been proposed including Statistical parity~\cite{feldman2015certifying}, Equalopp and Equalodds~\cite{hardt2016equality}.
Statistical parity requires that the probability of positive output of different groups is exactly equal, ignoring the label distribution of the test set itself.
% EqualOpp measures bias by comparing true positive rate between different groups. Comparison results on \textbf{Equalopp} are reported in the Appendix.
However, the fairness of positive and negative outputs is equally important, such as \emph{blonde} (Pos.) and \emph{non-blonde} (Neg.) in hair color recognition. \textbf{Equalodds} comprehensively considers fairness on Pos. and Neg. as follows:
\begin{equation}
\frac{1}{|T|} \sum_{t} \left|\operatorname{Pr}_{b^{0}}(\tilde{T}=1 \mid T=t)-\operatorname{Pr}_{b^{1}}(\tilde{T}=1 \mid T=t)\right|
\end{equation}
where $T$ denotes target labels, $\tilde{T}$ denotes target outputs, and $b^{0}$ and $b^{1}$ represents different groups in terms of bias attributes such as \emph{male} and \emph{female}.

Early research on fairness measured accuracy using a test set with the same distribution as the training set, known as \textbf{Bias Accuracy}. However, it was discovered that models relying solely on bias features (\emph{e.g.}, gender) could achieve high Bias Accuracy on the target task (\emph{e.g.}, nurse) due to the test set being biased towards majority samples. While the goal of fairness is to guarantee performance across different groups of samples especially the minority, \textbf{Fair Accuracy} solves this problem by resampling to construct the fair test set with an equal number of samples for each target and bias pair. Fair Accuracy has become the preferred accuracy metric in recent fairness studies~\cite{jung2021fair}.

\myparagraph{Datasets.} We evaluate the debiasing performance of \textit{Active Shortcut Debiasing} on \textbf{CelebA}~\cite{liu2015deep} and \textbf{UTKFace}~\cite{zhang2017age}. CelebA annotates 40 binary attributes including two social attributes: \emph{Male} and \emph{Young}. We set Male (\emph{m}) and Young (\emph{y}) as bias attributes, and select Attractive (\emph{a}), Blonde (\emph{bl}), Big Nose (\emph{bn}), and Pale Skin (\emph{ps}) as target attributes. For UTKFace, we test the Ethnicity(\emph{e}) debiasing performance (with Male(\emph{m}) as target attributes) by constructing a biased training set. We use the original (relatively balanced) test set when testing.
We also test more general (beyond social bias) bias mitigation on the \textbf{Dogs\&Cats}~\cite{parkhi2012cats} and \textbf{BiasedMNIST}~\cite{bahng2020learning}. All the reported results are the averaged scores over three independent runs.

% For construct biased dataset from UTKFace, we truncate a portion of data to force the correlation $Pr(T|B)$ between target and bias attributes to be 0.9. 
% For unbiased evaluation of the accuracy and fairness, the test set was constructed to have same number of samples for each target and each bias on both CelebA and UTKFace.

\myparagraph{Baselines.} We compare Active Shortcut Debiasing (\textit{Active SD}) against baselines such as: (1) Training DNNs using cross-entropy loss without any debiasing technique (referred as \textit{Vanilla)}, (2) Adding a regularization term regarding fairness constraints in the model optimization objective, including \textit{AdvDebias}~\cite{wang2019balanced}, \textit{LNL}~\cite{kim2019learning}, \textit{End}~\cite{tartaglione2021end}, and \textit{MFD}~\cite{jung2021fair}, and (3) Controlling the contribution of bias features to the target task to achieve fair generalization, including \textit{DI}~\cite{wang2020towards}, \textit{RNF}~\cite{du2021fairness}, and \textit{FSCL}~\cite{park2022fair}.

\myparagraph{Implementation details.} 
We use \textbf{ResNet-18}~\cite{he2016deep} as the backbone network, and we initialize ResNet-18 with pretrained parameters. The vector dimension of shortcut features is set to 100. For all baselines, we randomly sample data with a batch size of 128 and use the Adam optimizer with a learning rate of 1e-3. To demonstrate the robustness of our method to model structures, we also use VGG-16~\cite{Simonyan15} as backbone network.

% \input{tables/table3}

% \vspace{-20pt}
\subsection{Debiasing Performance Comparison}
Table~\ref{table:celeba} shows the Bias accuracy, Fair accuracy, and the model bias (EqualOdds) of models on diverse combinations of target and bias attributes of CelebA. Averaging over all target and bias combinations, \emph{Vanilla} records the most severe model bias because it is optimized to capture the statistical properties of training data without any restrictions. Notably, our \emph{Active Shortcut Debiasing} (abbreviated as \textit{Active SD}) significantly outperforms previous methods in terms of both model bias and accuracy.
% 我们方法所表现的最优公平性验证了我们使用代理去偏见防止目标任务学习和去偏见之间竞争的motivation，实现了去偏见不被目标任务的学习所干扰。
% The fairness-accuracy compatibility validates our motivation that utilizes shortcut features to prevent the zero-sum game between target task learning and debiasing, which results in a win-win for our method in terms of accuracy and fairness.
Our method’s state-of-the-art performance in fairness confirms our motivation that using shortcut features can avoid the challenges posed by target task learning to debiasing. Moreover, the accuracy comparison between \emph{Vanilla} and \textit{Active SD} shows that there is no degradation in the accuracy of our method caused by an increase in fairness.
While the fairness in all other debiasing methods is more or less hindered by the learning of the target task. Methods such as \textit{AdvDebias}, \textit{LNL}, \textit{END} and \textit{MFD}, which aim to remove bias features from representation, show limited debiasing performance. 
% This suggests that the zero-sum game between target task learning and debiasing limits not only the accuracy but also the performance of debiasing.
% This suggests that the trade-off between target task learning and debiasing not only limits accuracy but also debiasing performance. 
Other methods, such as \textit{DI} and \textit{FSCL}, address this issue by controlling the contribution of bias attributes to achieve indirect debiasing. These methods have achieved accuracy and model bias second only to ours. It well validates the effectiveness of Shortcut Debiasing to resolve the incompatibility between the target task and the debiasing operation.

% Further, using T=$a$ and B=$m$ as an example, Figure~\ref{fig:tradeoff} illustrates the accuracy and debiasing performance of various methods under different parameters, such as regularization weights. Our method is positioned in the upper left corner of Figure~\ref{fig:tradeoff}, indicating that it significantly outperforms all state-of-the-art methods in terms of both accuracy and fairness in all settings. 

% We conjecture this is because \textit{DI} and \textit{FSCL} implicitly avoid competition between target tasks and debiasing.
% The standard deviations and comparison results for another model bias are provided in the appendix.

In addition to gender and age debiasing, we also compare ethnicity bias removal results on UTKFace in Table~\ref{table:utkface}.
Observations on ethnicity debiasing are consistent with those on gender and age debiasing on CelebA. This shows that the effectiveness of our method is independent of the bias attribute and is effective for the three commonly used sensitive social attributes.
% Table~\ref{table:utkface} summarizes the performance for different methods on UTKFace. The consistent observations with the above CelebA debiasing evaluation include: (1) \textit{Vanilla} records the most severe model bias. (2) Our \textit{Active PD} outperforms the previous methods in both model bias and accuracy (3) \textit{DI} performs second best in both model bias and accuracy due to \textit{DI} trains separate target task classifiers for each group. New observations include: For the case where the target attribute and the bias attribute are interchanged (T=\textit{e}/B=\textit{g} and T=\textit{g}/B=\textit{e}), our method can significantly eliminate the model bias, which shows that the debiasing ability of our method is independent of the setting of target attribute and bias attribute.

% \begin{figure}[t!] 
%     \centering
%     \includegraphics[width=0.45\textwidth]{figs/tradeoff.png}
%     \vspace{-8pt}
%     \caption{The fairness-accuracy curve comparison of all methods in different parameters.}
%      % \vspace{-5pt}
%     \label{fig:tradeoff}
% \end{figure}
\begin{table}[t!]

\caption{\textbf{The ethnicity debiasing performance on UTKFace.} The numbers in brackets note how much they are relative change (in $\%$) from vanilla.}
% \vspace{-5pt}

\centering
\resizebox{0.47\textwidth}{!}{
\begin{tabular}{c|ccc}
\toprule

\multirow{2}{*}{Method } 
& \multirow{2}{*}{Bias Acc. ($\uparrow$)}  & \multirow{2}{*}{Fair Acc. ($\uparrow$)} & \multirow{2}{*}{Bias ($\downarrow$)}
\\
\\ \cmidrule[0.5pt]{1-4} \morecmidrules\cmidrule[0.5pt]{1-4}
\textit{Vanilla} &88.78 &  88.53  & 15.70    \\ \cmidrule[0.5pt]{1-4}
\textit{AdvDebias} & 75.28 (15.20\textcolor{red}{$\downarrow$})& 74.15 (16.24\textcolor{red}{$\downarrow$}) & 27.94 (77.96\textcolor{red}{$\uparrow$})    \\ 
\textit{LNL} & 87.93 (0.95\textcolor{red}{$\downarrow$}) & 87.31 (1.37\textcolor{red}{$\downarrow$}) & 19.13 (21.84\textcolor{red}{$\uparrow$})\\ 
\textit{EnD} & 89.45 (0.75\textcolor{green}{$\uparrow$}) & 89.07 (0.60\textcolor{green}{$\uparrow$}) & 14.29 (8.98\textcolor{green}{$\downarrow$})  \\ 
\textit{MFD} & 88.23 (0.61\textcolor{red}{$\downarrow$}) & 87.75 (0.88\textcolor{red}{$\downarrow$}) & 13.74 (12.48\textcolor{green}{$\downarrow$}) \\ %\cmidrule[0.5pt]{1-6}
\textit{DI} & 90.04 (1.41\textcolor{green}{$\uparrow$}) & 89.81 (1.44\textcolor{green}{$\uparrow$}) & 1.62 (89.68\textcolor{green}{$\downarrow$}) \\ 
\textit{RNF} & 89.11 (1.13\textcolor{green}{$\uparrow$}) & 88.99 (0.51\textcolor{green}{$\uparrow$}) & 13.75 (12.42\textcolor{green}{$\downarrow$})  \\ 
\textit{FSCL} & 87.09 (1.90\textcolor{red}{$\downarrow$}) & 87.03 (1.69\textcolor{red}{$\downarrow$}) & 1.71 (89.10\textcolor{green}{$\downarrow$})  \\ \cmidrule[0.5pt]{1-4}
% \textit{FSCL} & 11.5 & 79.1 && 13.0 & 79.1  \\ 
\textit{Active SD} & \textbf{91.06 (2.56\textcolor{green}{$\uparrow$})} & \textbf{91.10 (2.90\textcolor{green}{$\uparrow$})} & \textbf{0.96 (93.88\textcolor{green}{$\downarrow$})}     \\ \bottomrule
\end{tabular}
}

\vspace{-6pt}

\label{table:utkface}
\end{table}

% \vspace{-10pt}
\subsection{Controlled Experiments in Various Data Bias} 
There are many cases of data imbalance in the real world. To assess the effectiveness and resilience of our method in handling these imbalances, we created several datasets with different imbalance ratios using UTKFace. For more details, we truncated a portion of data to force the correlation $Pr(T|B)$ between target(male) and bias(ethnicity) attributes to range from 0.3 to 0.9, simulating varying levels of bias.

\begin{figure}[t!] 
    \centering
    \includegraphics[width=0.48\textwidth]{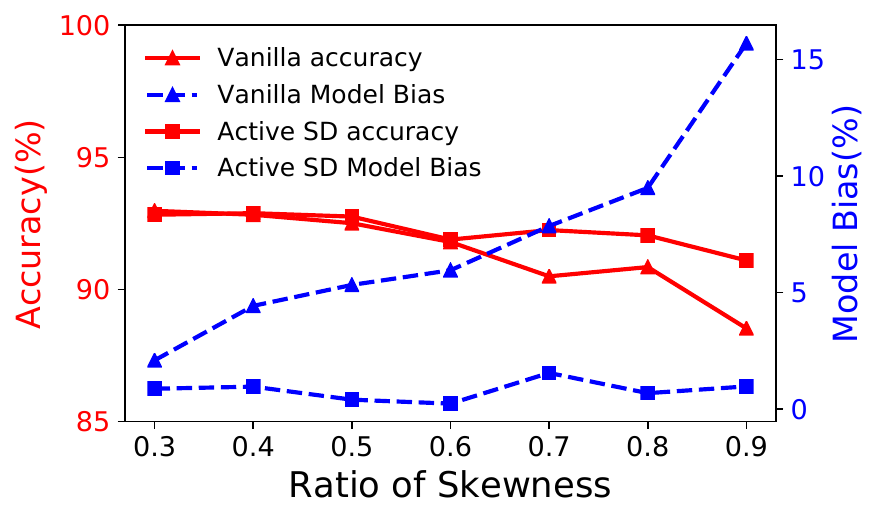}
    % \vspace{-5pt}
    \caption{\textbf{The Fair Accuracy (in $\%$, $\uparrow$) and model bias (Equalodds, $\downarrow$) of Active Shortcut Debiasing on UTKFace with varying ratio of $Pr$(\emph{male}$|$\emph{ethnicity}).}
    }
    \label{fig:databias}
    % \vspace{-5pt}
\end{figure}

In Figure~\ref{fig:databias}, we show Fair Accuracy and Model bias for vanilla and our method across different $Pr(T|B)$. 
Our method consistently outperforms vanilla in both accuracy and fairness, regardless of the intensity of data bias $Pr(T|B)$. Furthermore, as $Pr(T|B)$ increases, the accuracy gap between our method and vanilla widens. This demonstrates that our method is robust to diverse data bias scenarios and can yield greater benefits in more severely biased situations.

\subsection{The Effectiveness of Shortcut Effect Enhancement} 
We qualitatively validate the effectiveness of the shortcut effect enhancement module. This module optimizes the model to obtain sufficient biased information solely from shortcut features.

% This module optimizes shortcut features and models to actively make the target task depend on shortcut features, allowing the model to obtain sufficient biased information solely from these features.

\begin{figure}[t] 
    \centering
    \includegraphics[width=0.49\textwidth]{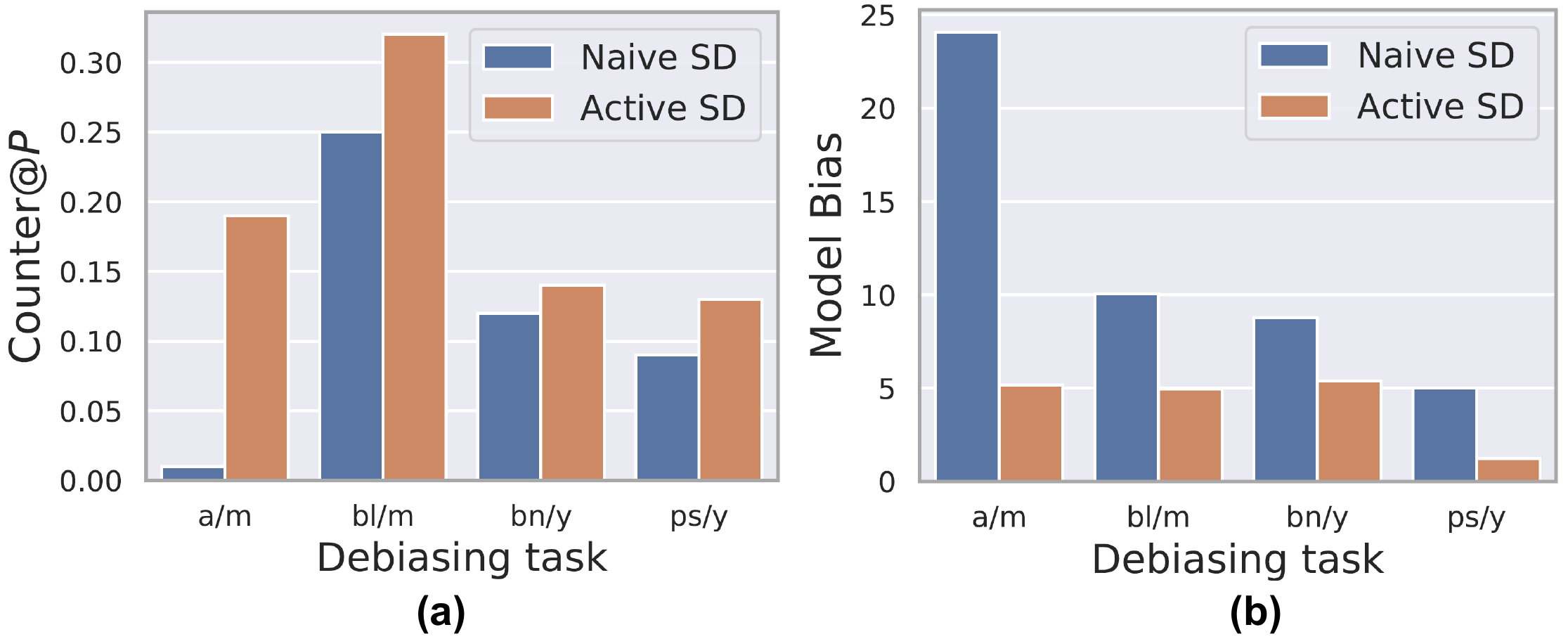}
    % \vspace{-20pt}
    \caption{The effectiveness of shortcut effect enhancement on CelebA. }
    % It clearly shows that as \emph{Counter@P} is improved by the \textit{shortcut effect enhancement} module in \textit{Active Shortcut Debiasing} (left), the debiasing effect is further improved (right).
    \label{fig:ablation}
    \vspace{-10pt}
\end{figure}

To validate, we use \emph{Counter@P}, as described in described in Eqn.~\ref{counterp}, to measure how dependent the model is on shortcut features in both \emph{Naive} and \emph{Active Shortcut Debiasing}. Figure~\ref{fig:ablation}(a) shows the \emph{Counter@P} values for both methods across four debiasing tasks on CelebA. The results demonstrate that \emph{Active Shortcut Debiasing} significantly increases the dependence of the target task on shortcut features (\emph{i.e}., higher \emph{Counter@P} values in all debiasing tasks). We also measure model bias to assess the dependence of the model on real bias features in both Shortcut Debiasing methods, as shown in Figure~\ref{fig:ablation}(b). Taken together, these plots reveal that as the model’s dependence on shortcut features increases, its reliance on real bias features decreases. This suggests that when our active shortcut strategy enables the target task to obtain sufficient biased information from shortcut features, the model no longer needs to use bias features.

\subsection{Qualitative Analysis with t-SNE Visualization}
\begin{figure}[t] 
    \centering
    \includegraphics[width=0.47\textwidth]{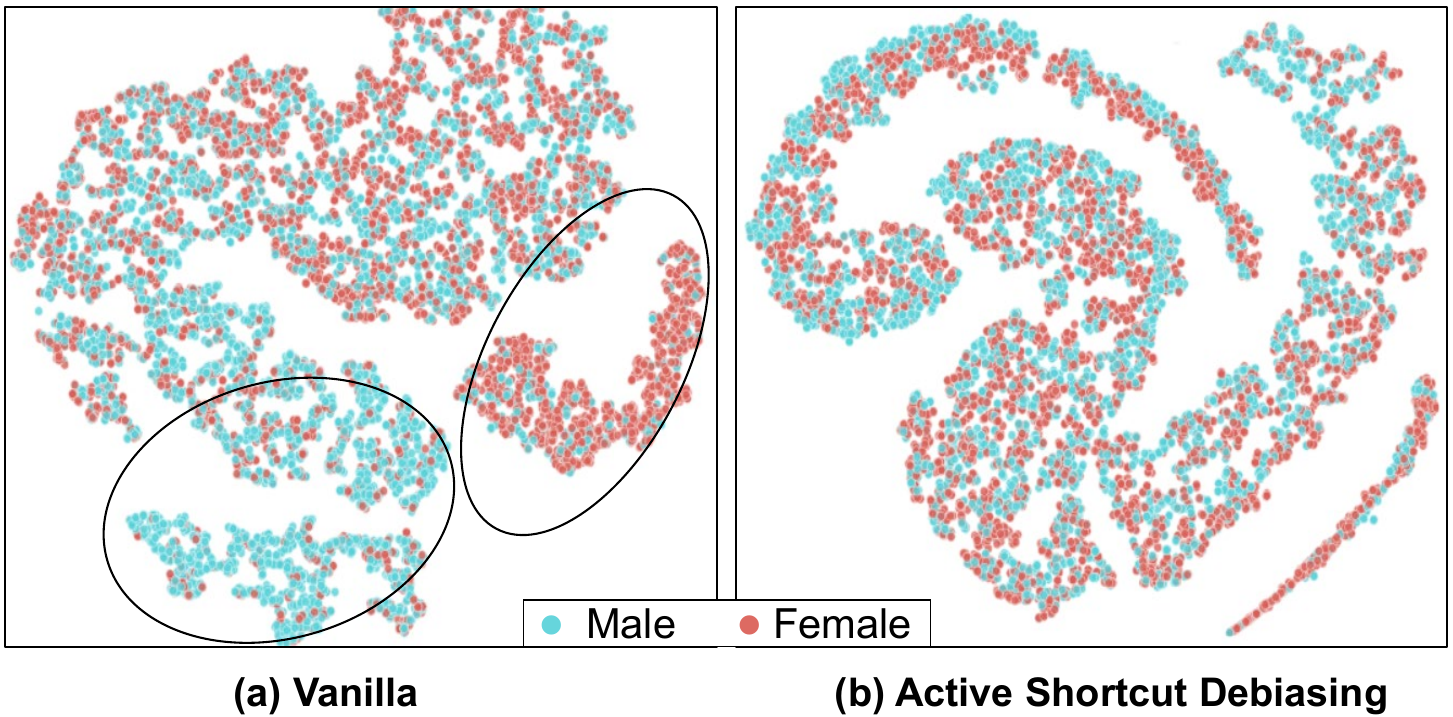}
    % \vspace{-5pt}
    \caption{\textbf{Qualitative comparison using t-SNE visualizations.} 
    }
    \label{fig:tsne}
    % \vspace{-5pt}
\end{figure}

We visualize t-SNE embeddings (Encoder $f(\cdot)$) of model trained with \emph{Vanilla} and \emph{Active Shortcut Debiasing} on CelebA.
The results for target =\textit{attractive} and bias =\textit{male} are presented in Figure~\ref{fig:tsne}(a) and (b). The points in the figure are divided into two groups based on \emph{gender} and are represented by different colors. 

In the \emph{Vanilla} model, the representation exhibits separability for the two bias attributes, particularly in the oval region shown in Figure~\ref{fig:tsne}(a), indicating that the model learns bias features during target task learning. In contrast, the representation in \emph{Active Shortcut Debiasing} cannot be divided by bias attribute, meaning that our method does not learn bias features from the data. This visualization demonstrates that our method effectively mitigates discrepancies between different groups.

\begin{figure}[t] 
    \centering
    \includegraphics[width=0.47\textwidth]{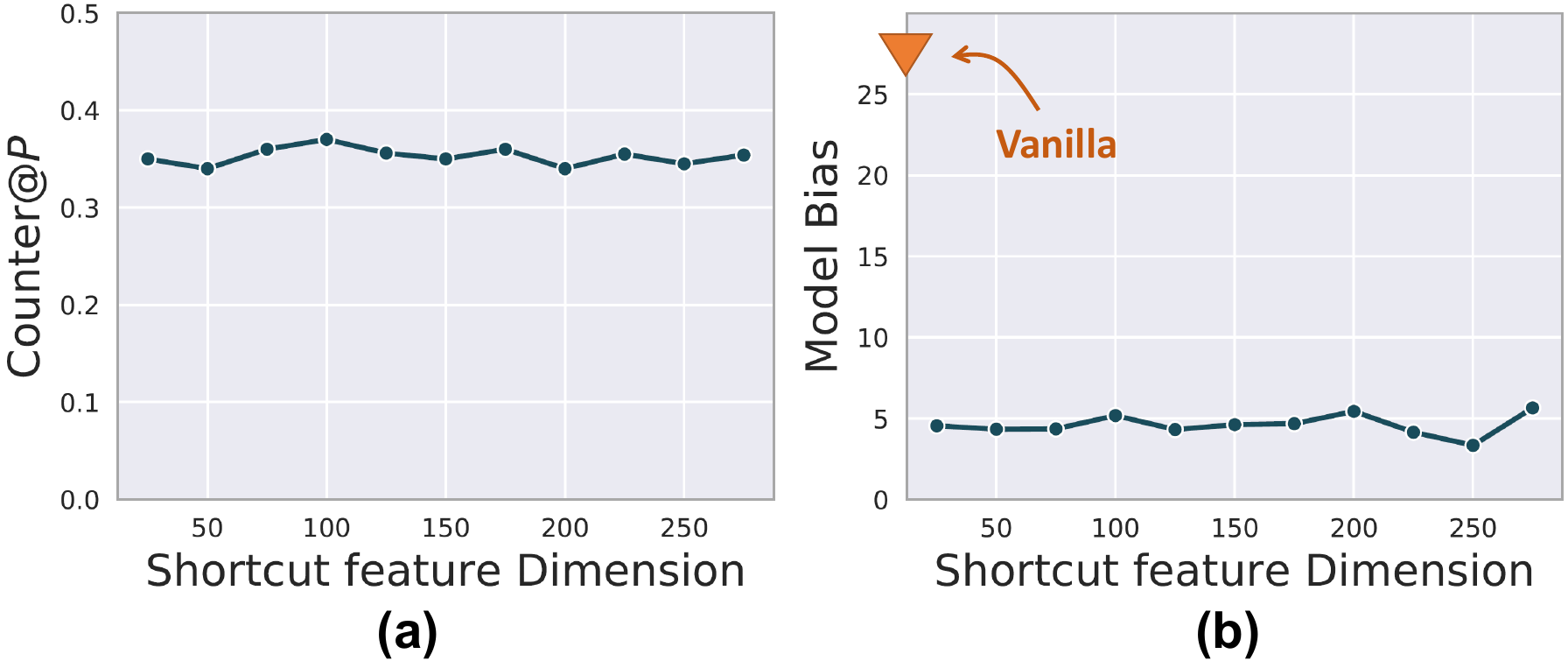}
    % \vspace{-5pt}
    \caption{Effect of the dimension of shortcut features.} 
    % \vspace{-5pt}
    \label{fig:dim}
\end{figure}

\subsection{Shortcut Feature Dimension Sensitivity}
We conduct a parameter sensitivity analysis on the dimension of shortcut features using the CelebA dataset. With target =\textit{attractive} and bias =\textit{male}, we ran \emph{Active Shortcut Debiasing} with various shortcut feature dimension settings and reported the results in Figure~\ref{fig:dim}, which includes both \emph{Counter@P} (a) and debiasing performance (b). Results indicate that the choice of shortcut feature dimension has no significant impact on either shortcut effect (\emph{Counter@P}) or debiasing. This suggests that our method can achieve effective debiasing using lower-dimensional shortcut features, resulting in lower computational costs.

\subsection{Robustness to Network Architectures}
To evaluate the robustness of \emph{Active Shortcut Debiasing} to different backbones, we report its performance using both ResNet-18 and VGG-16 as backbones in Table~\ref{table:modelArch}. The results demonstrate that our method performs similarly on both VGG and ResNet architectures, two mainstream network architectures, indicating that our method is not sensitive to the model architectures.

\begin{table}[t!]

   \caption{Effect of backbone architectures in Active Shortcut Debiasing on CelebA dataset.}
       % \vspace{-5pt}

\resizebox{0.47\textwidth}{!}{
\begin{tabular}{ccccc} \toprule
Method & Backbone & BiasAcc. ($\uparrow$) & FairAcc. ($\uparrow$) & Bias ($\downarrow$) \\ \cmidrule[0.5pt]{1-5} \morecmidrules\cmidrule[0.5pt]{1-5}
\multirow{2}{*}{\textit{Vanilla}} & \textit{Resnet} & 81.79$_{\pm 0.2}$ & 76.72$_{\pm 0.2}$& 26.24$_{\pm 0.5}$ \\ \cmidrule[0.3pt]{2-5}
 & \textit{VGG} & 81.83$_{\pm 0.2}$ & 77.35$_{\pm 0.1}$& 24.18$_{\pm 0.4}$ \\ \cmidrule[0.3pt]{1-5}
\multirow{2}{*}{\textit{Active SD}} &  \textit{Resnet} & 81.43$_{\pm 0.1}$ & 80.54$_{\pm 0.2}$ & 5.18$_{\pm 0.4}$\\ \cmidrule[0.3pt]{2-5}
 & \textit{VGG}& 81.38$_{\pm 0.2}$ & 80.33$_{\pm 0.2}$ & 5.76$_{\pm 0.3}$\\ \bottomrule
\end{tabular}
}

% 	\caption
% 	{
% % 	\small	
% 		
% 	}
\label{table:modelArch}
\end{table}

\begin{table}[t!]

\caption{The Bias Accuracy (in $\%$, $\uparrow$), Fair Accuracy (in $\%$, $\uparrow$), and model bias (in $\%$, Equalodds, $\downarrow$) of models trained on Dogs$\&$Cats and BiasedMNIST.}
% \vspace{-5pt}

\centering
\resizebox{0.45\textwidth}{!}{
\begin{tabular}{c|ccccccc}
\toprule
\multirow{3}{*}{Method } &\multicolumn{3}{c}{Dogs$\&$Cats} && \multicolumn{3}{c}{BiasedMNIST}  \\ 

\cmidrule[0.5pt]{2-4} \cmidrule[0.5pt]{6-8}  
& \multirow{2}{*}{BiasAcc.}  & \multirow{2}{*}{FairAcc.} & \multirow{2}{*}{Bias} && 
\multirow{2}{*}{BiasAcc.}  & \multirow{2}{*}{FairAcc.} & \multirow{2}{*}{Bias} 
\\
\\ \cmidrule[0.5pt]{1-8} \morecmidrules\cmidrule[0.5pt]{1-8}
% EqualizedOdds_max-min
\textit{Vanilla} &90.15& 91.64 & 14.29 &&90.66&90.60 & 28.21 \\ \cmidrule[0.5pt]{1-8}
\textit{AdvDebias} &89.37& 89.54 & 9.51 &&92.88& 93.13 & 12.21  \\ 
\textit{LNL} &90.19&  91.31 & 9.22 &&91.60& 91.51 &  19.12 \\ 
\textit{EnD} &91.44&  91.62 & 14.64 &&91.40&91.47 & 21.35 \\
\textit{MFD} &91.97&  91.62 & 9.71 &&90.50&90.58 & 24.82 \\ 
\textit{DI} &92.53&  92.75 & 6.11 &&93.22&93.24 & 11.33  \\
\textit{RNF} &90.01& 90.53 & 14.29 &&89.94& 89.23 & 27.10 \\ 
\textit{FSCL} &92.13& 92.97 & 6.69 &&94.12& 94.40 & 12.36  \\%\cmidrule[0.5pt]{1-18}
\cmidrule[0.5pt]{1-8} 
% \textit{FSCL} & 11.5 & 79.1 && 13.0 & 79.1 && 7.0 & 82.1 && 6.4 & 83.8 && 3.8 & 82.7 && 1.8  & 82.0     \\ 
\textit{Active SD} & \textbf{94.48}&\textbf{94.87} & \textbf{4.23} && \textbf{96.38} &\textbf{96.40} & \textbf{7.02}  \\ \bottomrule
\end{tabular}
}

\vspace{-5pt}
\label{table:generalBias}
\end{table}
\subsection{Extensibility to General Bias Mitigation}
To evaluate the effectiveness of our proposed methods in addressing general bias types, we conducted experiments on the Dogs\&Cats dataset with color bias. We also assessed the debiasing performance of our method on multi-classification tasks by performing a ten-digit classification task on the BiasedMNIST dataset, with the goal of eliminating the corresponding ten color biases. The results, shown in Table~\ref{table:generalBias}, demonstrate that our method best eliminates both color biases, indicating their generalizability to various types of bias. Noted that debiasing approaches such as \emph{DI} and \emph{FSCL} exhibit higher Bias accuracy than \emph{Vanilla}, as fairness improves accuracy performance on a relatively balanced test set.

\section{Conclusion}
% 我们工作的目标是消除偏见的同时不破坏有用的目标任务信息.To this end,我们提出了对目标任务相关特征的因果干预来直接获得模型由目标信息作出的无偏预测,对于因果干预,我们使用了后门调整来实现.为了更有效的对偏见相关特征进行因果干预,我们提出了特征代理的方法使用人工特征代理模型中target task对偏见信息的应用.在多种数据集上的实验效果展示了我们方法significantly improves 之前的方法在准确率和去偏见上.

In this paper, we propose the use of shortcut features to replace the role of bias features in target task learning for visual debiasing. Our solution integrates the enhancement of shortcut effects and inference with the intervention feature, avoiding the use of bias features during training and shortcut features during testing. The introduction of shortcut features solves the problem of debiasing being hindered by target task learning in previous methods, and the experimental results demonstrate its effectiveness in improving fairness and accuracy.

\begin{acks}
This work is supported by the Beijing Natural Science Foundation (No.JQ20023).
\end{acks}

% \vspace{-1mm}
% \begin{acks}
% This work is supported by the National Key R\&D Program of China (Grant No. 2018AAA0100604), and the National Natural Science Foundation of China (Grant No. 61632004, 61832002, 61672518).
% \end{acks}


\bibliographystyle{ACM-Reference-Format}
\balance
\bibliography{main}

\end{document}